\documentclass[conference]{IEEEtran}
\IEEEoverridecommandlockouts
\usepackage{cite}
\usepackage{amsmath,amssymb,amsfonts}
\usepackage{algorithmic}
\usepackage{graphicx}
\usepackage{textcomp}
\usepackage{xcolor}
\def\BibTeX{{\rm B\kern-.05em{\sc i\kern-.025em b}\kern-.08em
    T\kern-.1667em\lower.7ex\hbox{E}\kern-.125emX}}

\usepackage{hyperref}

\usepackage{graphicx,url}
\usepackage{subcaption}
\usepackage{comment}
\usepackage{ifthen}
\usepackage{tikz}
\usepackage{caption}
\usepackage{xurl}
\usepackage{caption}

\begin{document}

\title{Are Explanations Helpful? A Comparative Analysis of Explainability Methods in Skin Lesion Classifiers
}

\author{\IEEEauthorblockN{Rosa Y. G. Paccotacya-Yanque$^{1,2}$, Alceu Bissoto$^{3}$, Sandra Avila$^{1}$}
\IEEEauthorblockA{$^{1}$Instituto de Computação, Universidade Estadual de Campinas (UNICAMP), Campinas, Brazil \\
    $^{2}$Department of Computer Science, Universidad Católica San Pablo, Arequipa, Peru \\
$^{3}$Department of Diabetes, Endocrinology, Nutritional Medicine and Metabolism UDEM, University of Bern, Switzerland \\
}}

\maketitle

\begin{abstract}
Deep Learning has shown outstanding results in computer vision tasks; healthcare is no exception. However, there is no straightforward way to expose the decision-making process of DL models. Good accuracy is not enough for skin cancer predictions. Understanding the model's behavior is crucial for clinical application and reliable outcomes. In this work, we identify desiderata for explanations in skin-lesion models. We analyzed seven methods, four based on pixel-attribution (Grad-CAM, Score-CAM, LIME, SHAP) and three on high-level concepts (ACE, ICE, CME), for a deep neural network trained on the International Skin Imaging Collaboration Archive. Our findings indicate that while these techniques reveal biases, there is room for improving the comprehensiveness of explanations to achieve transparency in skin-lesion models.
\end{abstract}


\section{Introduction}
Deep Neural Networks~(DNNs) have shown promising results across different tasks. However, there is a lingering black-box perception of DNNs, making it difficult to understand how and why they make these decisions~\cite{DLSurveyACM18-10.1145/3234150}. This lack of transparency is particularly concerning in critical domains such as healthcare and criminal justice, where decisions influenced by DNNs can impact human lives. To build trust and integrate intelligent systems into daily life, we must build ``transparent'' models that explain why they predict what they predict~\cite{gradcam-selvaraju2017grad}. eXplainable Artificial Intelligence (XAI) aims to elucidate~how models make decisions, facilitating the understanding of DNNs through two approaches: self-explanatory models, designed to be inherently interpretable, and post-hoc explanations, generated using external methods applied to trained models. XAI promotes fairness in decision-making, facilitates robustness, and assures that only meaningful variables contribute to the output~\cite{BARREDOARRIETA202082}.

Melanoma, a highly aggressive form of skin cancer, can be effectively treated if detected early~\cite{early-diagnosis-cancer}. DNNs have shown remarkable performance in classifying skin lesion images, often surpassing human experts~\cite{valle2020data,bissoto2023artifact, chaves2023evaluation,HAGGENMULLER2021202, Acnns24, MIR2024112,bissoto2023test}. 
To deploy these models in clinical settings and support specialists, it is crucial to understand their behavior and ensure their explanations align with medical knowledge. In this scenario, the main contributions of this paper are: 
\begin{itemize}
    \item We describe a set of properties that an explanation should accomplish for skin-lesion classifiers.
    \item We conduct a study on the current use of XAI in skin-lesion analysis.
    \item We compare seven post-hoc explanatory methods, four methods based on pixel attribution: Grad-CAM~\cite{gradcam-selvaraju2017grad}, Score-CAM~\cite{wang2020score}, LIME~\cite{Ribeiro-LIME10.1145/2939672.2939778}, and SHAP~\cite{shapley1953value}, and three methods based on concepts: ACE~\cite{ghorbani2019towardsACE}, ICE~\cite{zhang2021invertible}, and CME~\cite{Kazhdan2020}. 
    We qualitatively evaluate the explanations in the most crucial and public skin-lesion dataset: the International Skin Imaging Collaboration (ISIC) Archive~\cite{isicarchive}. 
\end{itemize}
  
\section{Related Work}
We conducted a review following the PRISMA guidelines~\cite{prisma} on works that use XAI for DNNs with dermoscopic images from January 2020 to December 2023 and identified 42 studies. Pixel-attribution methods were the most commonly employed form of explanation, used in nearly 76\% (32/42) of the reviewed articles, adopted techniques included CAM, Grad-CAM, LIME, SHAP, and Attention-based methods. 

Regarding the purpose of employing XAI, around 55\% (23/42) used XAI superficially: 18/23 as a sanity check to show that the classifier focuses on the lesion or parts of it without any further analysis or evaluation and 5/23 as a sanity check (pipeline) that included the XAI method in the model for self-explanation without further analysis. From all the selected articles, about 7\% (3/42) presented new XAI methods and showed their application in skin lesion models, while close to 14\% (7/42) improved previous XAI methods, e.g., improving the resolution of the saliency map~\cite{Shinde2021}, turning an ante-hoc XAI method into a post-hoc method~\cite{Yuksekgonul2022}. Nearly 21\% (9/42) presented a detailed analysis and evaluation of their results using XAI methods for skin lesion models. 

Furthermore, only 21\% (9/42) included quantitative metrics to evaluate the explanation. The most common quantitative way to assess pixel-attribution methods was to compare segmentation masks versus the saliency map, i.e., check how many important pixels are in the skin lesion.

\section{Methodology}
    \subsection{Dataset Selection and Models}
    
    The ISIC releases a challenge annually with various tasks and datasets. The ISIC~2018 Challenge~\cite{isic2018} included lesion segmentation, attribute detection, and disease classification. We used images from Task 3 for training models to classify skin lesions as melanoma or benign.
   
    We focus on studying an Inception-v4 because of its notable performance and popularity in this task~\cite{valle2020data}. The network is pre-trained on ImageNet and fine-tuned 
    with stochastic gradient descent (momentum $0.9$, weight decay $0.001$, learning rate $0.001$, which was adjusted using a plateau scheduler by monitoring the validation loss with patience of $10$ epochs, a reduction factor of $10$, and a minimum learning rate of $10^{-5}$). We conducted the training for a maximum of $100$ epochs, with early stopping based on validation loss (patience of $22$ epochs) and a batch size of $32$. Data augmentation techniques were applied to training, validation, and test sets. Over six runs, the \text{Inception-v4} model obtained $89.96\% \pm 0.52$ Area Under the Receiver Operating Characteristic Curve (ROC AUC).
   
    We required a dataset with annotated dermoscopic attributes and their image localization to evaluate the results from the visual explainability methods. Therefore, we selected the ISIC 2018 Task 2 dataset for its detailed annotation of features such as pigment network, negative network, streaks, milia-like cysts, and dots/globules. This facilitates a comparison between the visual explainability results and the annotated dermoscopic attributes in the images.

    \subsection{Explainability Methods}
    
    Pixel attribution methods provide heatmaps with pixels highlighted according to their importance for prediction, and concept-based methods provide human-understandable concepts.
    We selected four based on pixel-attribution: \text{Grad-CAM~\cite{gradcam-selvaraju2017grad}}, Score-CAM~\cite{wang2020score}, LIME~\cite{Ribeiro-LIME10.1145/2939672.2939778}, SHAP~\cite{shapley1953value}, and three methods based on  high-level concepts: ACE~\cite{ghorbani2019towardsACE}, ICE~\cite{zhang2021invertible}, and CME~\cite{Kazhdan2020}. The last two methods, CME and ICE, create a self-explanatory surrogate model from the inner representations of the target models. While the first six methods provide explanations using lesion images, CME requires external knowledge, such as metadata that serves as concepts, to generate explanations.
    We specify the hyperparameters used for each method on \href{https://github.com/RosePY/XAI-for-Skin-Lesion}{GitHub}.
  
    \subsection{Analysis and Evaluation of Results} 
    We assessed the quality of the explanations and identified three properties they should accomplish for understandable skin-lesion models:
    
    \textbf{Fidelity: }This property indicates how well the explanation approximates the behavior of the black-box model~\cite{molnar2019,Robnik2018,samek2019xaibook}. It is also known as faithfulness~\cite{schwalbe2021comprehensive} and accuracy~\cite{carvalhosurvey}. Depending on the XAI method type, it can be evaluated quantitatively by inserting and removing the most relevant features according to the explanation and monitoring the change in the prediction, or qualitatively by comparing obtained explanations from different techniques for predictions with the highest confidence. For surrogate models, the fidelity is given by the approximation quality of predictions. Currently, different metrics exist to measure this property~\cite{MIRONICOLAU2024104179}.

    \textbf{Meaningfulness: } This refers to how well users understand and find the explanations coherent~\cite{schwalbe2021comprehensive}. It is also referred to as comprehensibility~\cite{molnar2019,Robnik2018} and understandability~\cite{carvalhosurvey,samek2019xaibook}. A meaningful explanation is one that users can easily grasp and interpret in a way that aligns with their existing knowledge.

    \textbf{Effectiveness: }
    The explanation should be detailed enough to enable users to simulate the model's outcome and generate testable hypotheses~\cite{schwalbe2021comprehensive}. This property is also referred to as sufficiency~\cite{samek2019xaibook}.

    We will assess the last two properties, meaningfulness, and effectiveness, according to the audience's willingness to understand the model's inner workings.

\section{Results and Discussion}
\begin{figure}[h]
    \centering
    \includegraphics[trim={0cm 0cm 0cm 0cm},clip,width=0.45\textwidth]{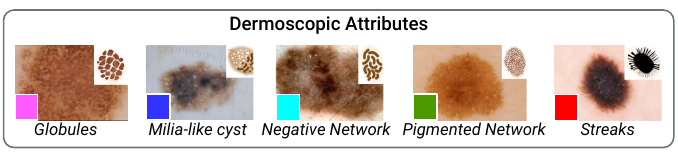}\\ \vspace{0.05cm}
    \includegraphics[trim={0cm 0cm 8cm 1.7cm},clip,width=0.45\textwidth]{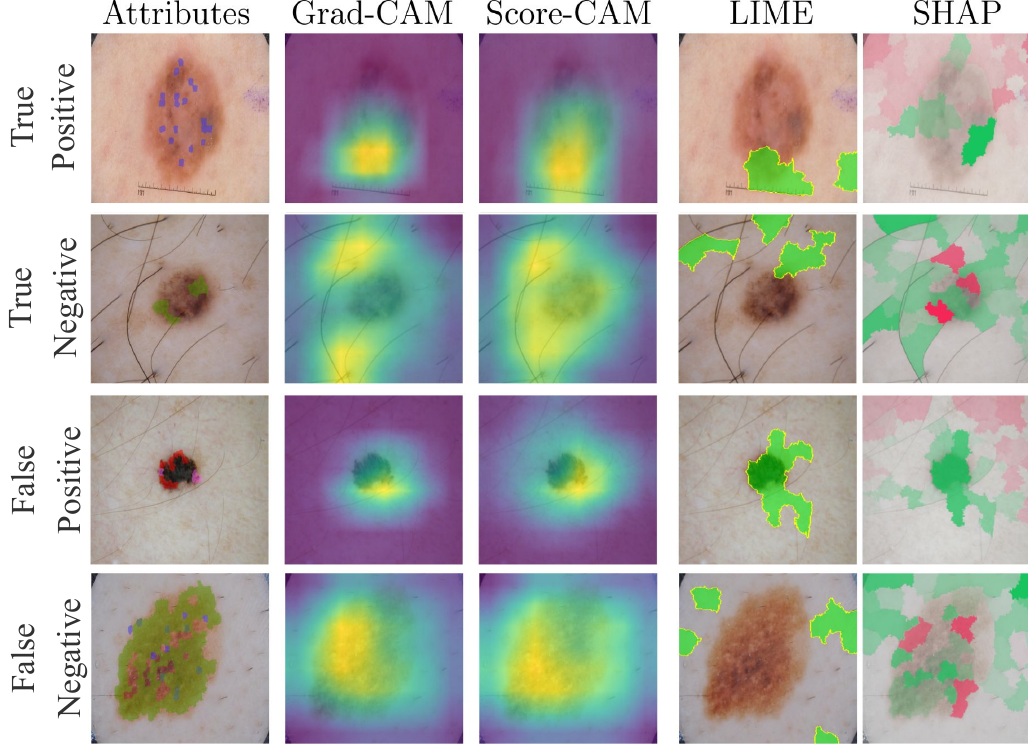}
    \caption{Pixel attribution results: \textcolor{yellow}{Yellow} indicates relevance for the prediction. In LIME, \textcolor{green}{green} highlights positive contributions. In SHAP, \textcolor{green}{green} pixels contribute positively, while \textcolor{red}{red} pixels contribute negatively.}
   \label{fig:pixel_attr}
\end{figure}

Fig.~\ref{fig:pixel_attr} shows four examples from the pixel-attribution methods for each model (each one presents a skin-lesion image with dermoscopic attributes, and the superimposed images) according to the correctness of the prediction: true positive,  true negative,  false positive, and false negative. The superimposed images are known as local explanations, as they explain only one prediction.\vspace{-0.2cm}

\begin{figure*}
\centering
    \begin{subfigure}[b]{\textwidth}
         \centering
         \includegraphics[trim={0.3cm 2.6cm 6.2cm 0cm},clip,width=0.72\textwidth]{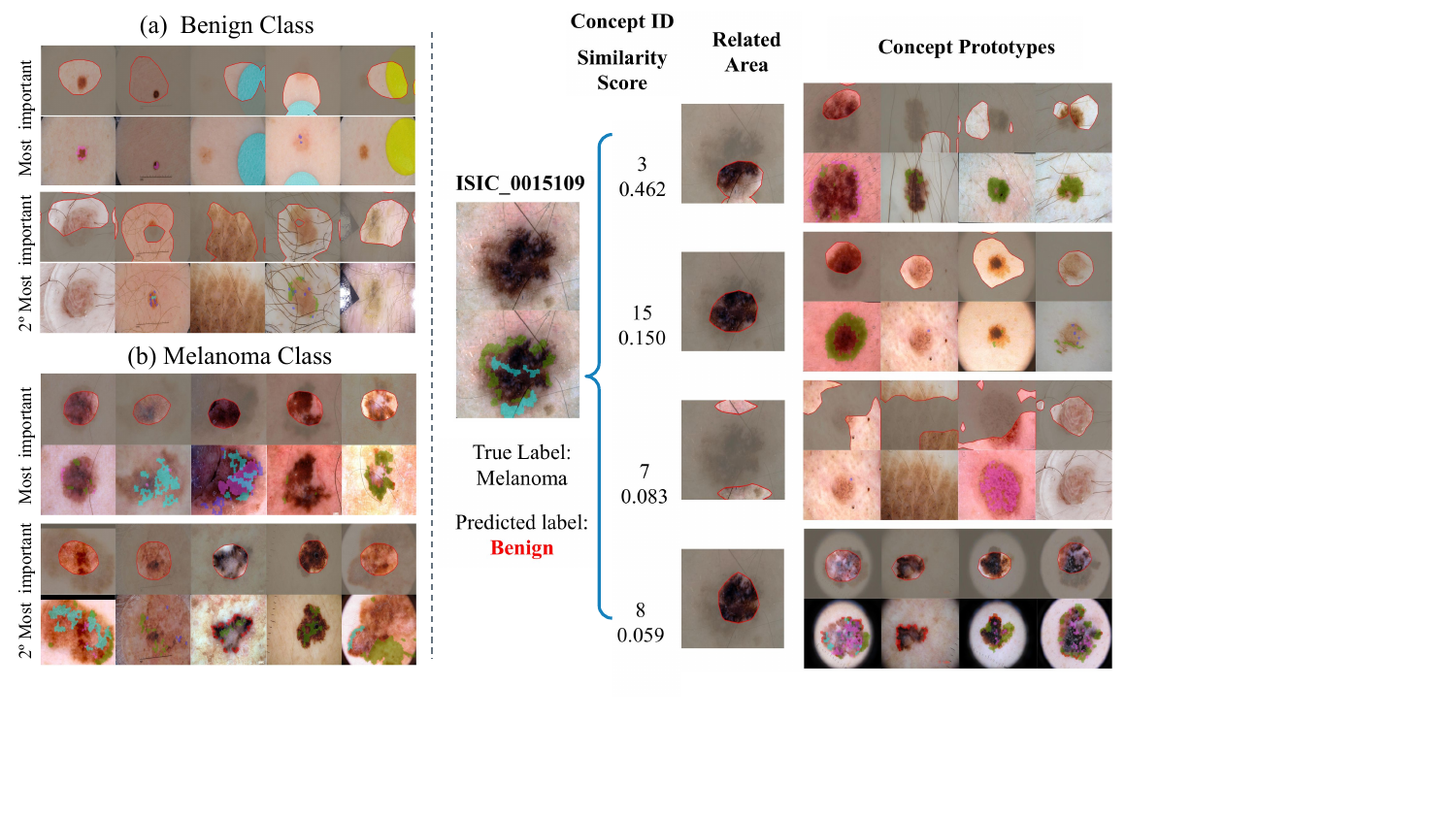}
         \caption{ICE Inception-v4 results -- Fidelity error: 11.83\%. Left: Two most important concepts for each class. Right: An example of an explanation for a false negative prediction, each area shows samples of the related concept.}
         \label{fig:ice}
    \end{subfigure}\\\vspace{0.5cm}     
    \begin{subfigure}[b]{0.34\textwidth}
        \begin{minipage}{\linewidth}
        \centering
        \includegraphics[trim={0cm 14.5cm 5.3cm 0cm},clip,width=0.95\textwidth]{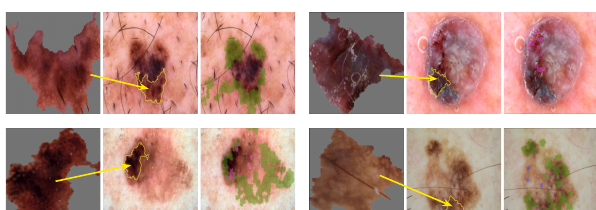}
        \caption{Most salient concept}
        \end{minipage}\\
                \begin{minipage}{\linewidth}
        \centering
        \includegraphics[trim={0cm 9.8cm 5.3cm 4.7cm},clip,width=0.95\textwidth]{images/ace_results_iv4.pdf}
        \caption{2nd most salient concept}
        \end{minipage}
         \caption{ACE results: Random examples of the two most important concepts in the last convolutional layer for the melanoma class. Each example shows the segment, its position in the image, and the corresponding dermoscopic attributes.}
         \label{fig:ace} 
    \end{subfigure}
        \hfill
    \begin{subfigure}[b]{0.64\textwidth}
         \centering
        \includegraphics[width=\textwidth]{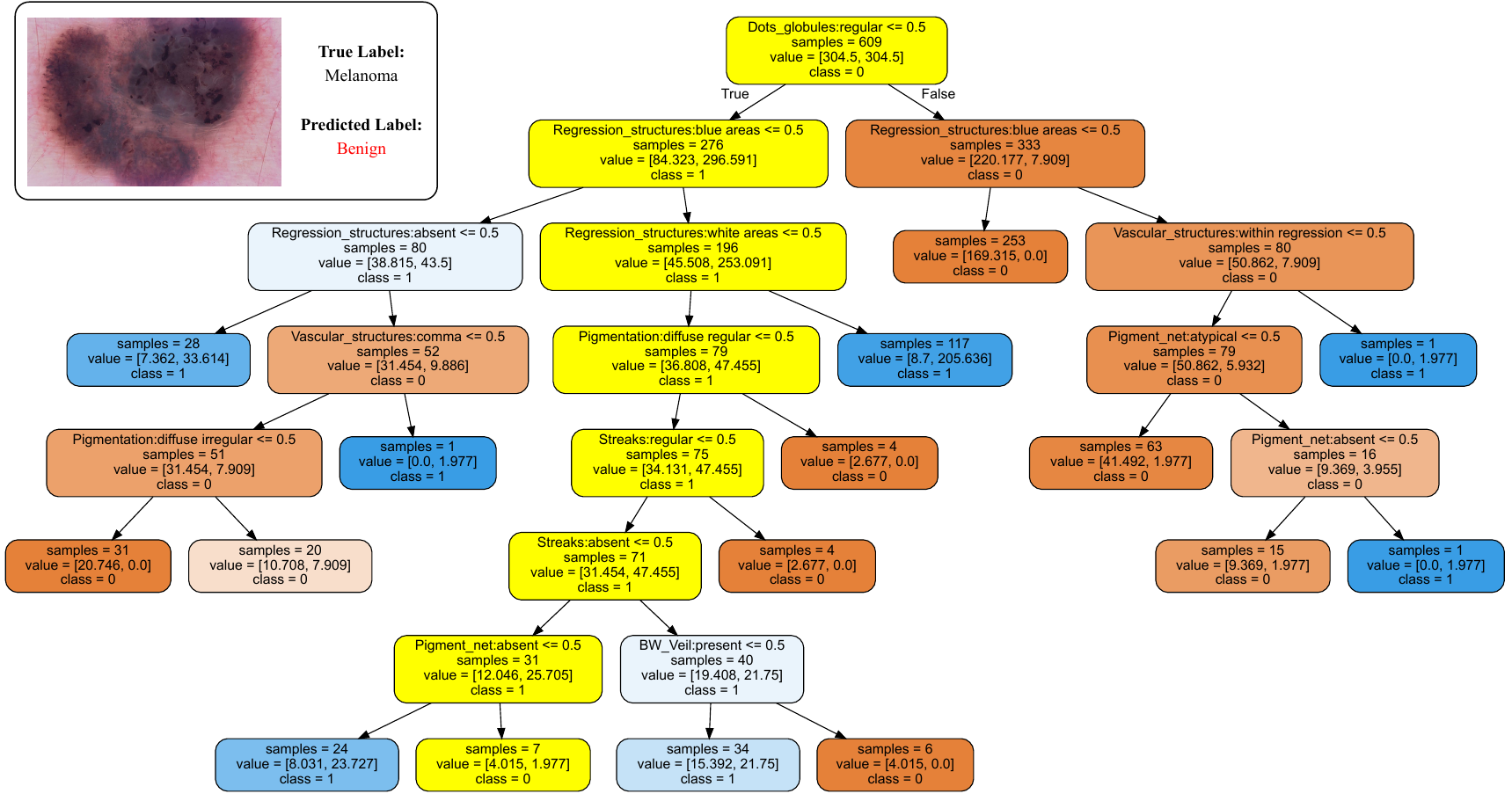}
         \caption{Decision tree for Inception-v4 using CME -- Fidelity: 0.88 AUC. The path explaining the prediction for the image in the upper left corner is highlighted in yellow.}
        \label{fig:cme} 
    \end{subfigure}
    \caption{Concept-based explanations results.}
    \label{fig:concept_exp}
\end{figure*}

In Fig.~\ref{fig:pixel_attr}, we can see most of the methods focus on spurious correlations, i.e., visible artifacts \cite{bissoto2019constructing,bissoto2020debiasing,bissoto2024even} introduced during the image acquisition process, e.g., patches, gel bubbles, and ruler marks. 
Here, the model bases its decision on the surrounding parts of the skin lesion and even on other artifacts, posing a risk to patients who may rely on the automatic diagnosis. Furthermore, these explanations are insufficient because they only say, ``the classifier predicts the melanoma class because there is a skin lesion''. However, they do not provide any further information to support the prediction.

Figs.~\ref{fig:ice},~\ref{fig:ace},~and~\ref{fig:cme} show the results for concept-attribution methods. ACE and ICE find the concepts and their localization in an unsupervised setting, and CME uses textual concepts without localizing them. These three methods provide global explanations (explain a class), but only ICE provides local ones (explain the prediction for a single input). 
For ACE (Fig.~\ref{fig:ace}), we found it hard to interpret the obtained concepts. At first sight, the most salient concept seems to be related to the pigmented network attribute. However, this attribute is also present in other concepts. 

For ICE (Fig.~\ref{fig:ice}), it is challenging to interpret the results for the melanoma class. We can see that the concept samples show lesion parts; sometimes, there is no corresponding dermoscopic attribute, or they are not the same across the samples. Since the concepts identified by ICE and ACE can fail to match the available attributes, the network may not base its decision on these dermoscopic attributes. Furthermore, since ICE uses activations and a threshold in the feature map to visualize concepts, the sample concepts are like those obtained with CAM methods, i.e., they show a big part of the lesion without indicating a specific part, making it difficult to contrast them with the dermoscopic attributes. Concept samples are, in general, very alike between them in color and texture.

Alternatively, CME uses model distillation to explain a neural network. First, it maps the layer's activations to concepts, and then, it gets the final output from the concept prediction. To train a Concept Predictor, we used Derm7pt dataset~\cite{derm7pt} that contains 870 dermoscopy images evaluated with the \text{7-point} criteria, which allowed us to have 25 concepts in tabular format. Fig.~\ref{fig:cme} shows the Inception-v4 distilled into a decision tree using CME. We can trace a path in a decision tree to understand what rules the model followed to make a particular prediction. Furthermore, CME allows to perform a concept intervention, meaning that a specialist can correct the concept predictions and see how this changes the model prediction. Also, if the distilled model produced by CME is retrained with concept intervention, the model's fidelity increases~notably. Although a decision tree is employed, other inherently interpretable models, such as logistic regression, could also be used, where the concept weights can be evaluated through odds ratios.

\begin{figure}[h]
    \centering
    \includegraphics[trim={0cm 0cm 0cm 0cm},clip,width=0.49\textwidth]{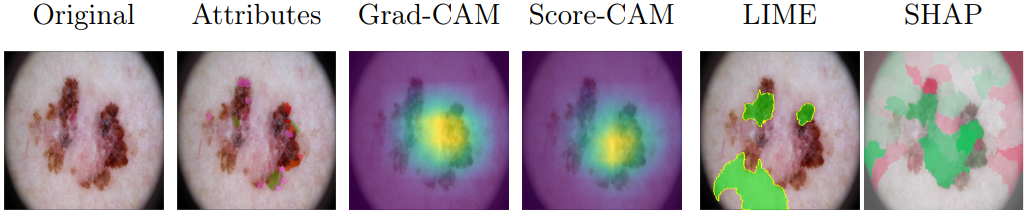}\\ \vspace{0.05cm}
    \includegraphics[trim={0cm 0cm 0cm 0cm},clip,width=0.49\textwidth]{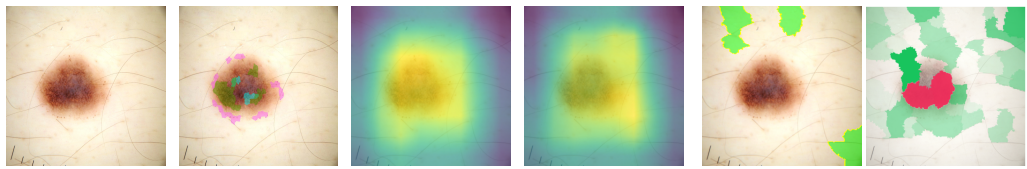}
    \caption{Saliency results for predictions with high confidence to test fidelity. First row: Melanoma class, second row: Benign class.}
   \label{fig:fidelity}
\end{figure}

Regarding our desiderata for explanations, we assessed \textit{fidelity}, which measures how much the explanations reflect the real model. We performed a qualitative evaluation following the methodology in~\cite{Sun2020} comparing explanations for correctly predicted images. Ideally, if the methods accurately explain the model behavior, the explanations across all methods will be similar. However, this is not the case, as shown in Fig.~\ref{fig:fidelity}. Some of the used methods are built to have the highest fidelity possible, such as ICE and CME, which are based on factorization and distillation techniques, respectively. Fidelity can be quantified using built-in metrics specific to each method. For ICE, which uses non-negative matrix factorization, the fidelity measured by the relative error was 11.83\%. For CME, using decision trees, the fidelity was 0.88 ROC AUC. Thus, around 10\% is getting an incorrect explanation, a considerable percentage for a crucial domain as it is identifying~cancer. 

Concerning the \textit{meaningfulness} of explanations, pixel-attribution methods might be more intuitive and meaningful for data scientists since highlighting important regions offers a quick and accessible method for interpreting where the model is focusing its attention, but for clinicians looking at a group of pixels could not be coherent at all. Concept-based explanations provide higher-level information than pixel-attribution ones. CME ensures human comprehension by associating machine parameters with demoscopic attributes. However, the comprehensibility of the other concepts-based methods (ICE and ACE) that find concepts in an unsupervised setting, i.e. by extracting concepts from the images used to train the network, requires domain knowledge and can be time-consuming. Also, there is no guarantee that the concepts found are related to human concepts since the examples of concepts are determined and limited to the superpixel segmentation algorithm (ACE) and the threshold to limit the attention map (ICE).

About the \textit{effectiveness} of explanations, which ensures that the explanation is complete, and users (data scientists, clinicians, and patients) could use it to form and test their own hypotheses. Pixel-attribution methods do not provide enough information to understand why a prediction is being made. They highlight the relevant parts but do not indicate how the model uses them to lead to a particular prediction or what the DNN model does with those specific parts of the images. Therefore, it is not possible to generate a hypothetical set of rules to simulate the model's outcome. On the other hand, concept-attribution methods provide more information about how the model uses the concepts to get the model output, e.g., Fig.~\ref{fig:ice} on their second columns shows how the different concepts influenced a melanoma lesion to be predicted as~benign.

The final two desired properties are inherently subjective and were evaluated primarily from the perspective of data scientists. It is important to note that assessments of these properties may differ among other stakeholders, including patients, physicians, and dermatologists. 

Pixel-attribution methods (Grad-CAM, Score-CAM, LIME, and SHAP) help to ensure that the models work appropriately by checking the presence of biases and spurious correlations. These methods produce local explanations, revealing the model's behavior for each image. This could assist clinicians in tailoring their explanations to individual patients, enhancing patient engagement and understanding. In concept-based methods such as ACE and ICE, the produced global explanations identify the patterns models use to predict a particular class. These methods find important features, biases, and spurious correlations faster than pixel-attribution methods. For the benign class, for example, in Fig.~\ref{fig:ice}, we can note how the most important concepts are related to artifacts such as color patches and skin hair. Nevertheless, comprehending the concepts these methods provide as an explanation for the melanoma class is difficult and requires domain knowledge. At first sight, CME is a good option for understanding DNNs, given that it uses known high-level concepts to justify the DNN's decision. On the other hand, even a fidelity of over 88\% leaves a great chance of doubts about whether to trust the explanation. Furthermore, CME is forcing the DNN's representations to be mapped to the provided human concepts; this could limit finding new concepts or spurious correlations, making it easier to believe that the network is performing desirably mistakenly. 

In summary, even when pixel-attribution methods provide instance-level explanations that are relatable for patients, offering intuitive and easily understandable insights into how the model makes decisions, they may not be sufficient on their own in the context of melanoma classification, where explainability is crucial for building user trust and acceptance among clinicians and patients, since these methods do not justify how the DNNs are using the relevant pixels or superpixels. To address this, it is necessary to complement pixel attribution methods with additional explanation techniques, such as concept-based explanations that provide higher-level insights and contextual~information.

\section{Conclusion}
We compared seven state-of-the-art explainability techniques, which use different approaches. However, none provide a sufficient explanation to fully understand the DNN's decisions. Combining methods can help compensate for the individual limitations of each technique, providing a more comprehensive understanding of the network's behavior. Each method has its advantages and disadvantages. Current XAI methods should not be expected to accomplish all explainability goals.  For instance, while identifying biases and spurious correlations is crucial, it may not directly aid decision-making or enhance trust. Consequently, caution is advised when utilizing and developing XAI methods, with clear consideration of their intended goals and application contexts.
Future directions include a deep study of how physicians perceive and interpret each explanation, and expanding our study to use other architectures and datasets~\cite{ricci2023dataset}.

\section{Ethics Statement}
\label{sec:ethics}

The datasets used for this work have an over-representation of lighter skin tones \cite{barros2023assessing,avila2023castacblog}. This bias has affected not only the DNN models used in this research but could also be perpetuated by the explanations. Also, having post-hoc explanations can lead to a false sense of understanding and confidence in AI models. XAI should not be seen as an escape from accountability but as a step towards a responsible deployment of AI~\cite{lima_faact22}. 

\section*{Acknowledgment}

R. Paccotacya-Yanque was funded by CNPq 155459/2019-8 and Bolsa Alumni from the Institute of Computing. S. Avila is funded by FAPESP 2023/12086-9, 2020/09838-0, 2013/08293-7, H.IAAC 01245.003479/2024-10, and CNPq 316489/2023-9. 

\bibliography{mybib.bib}{}
\bibliographystyle{plain}


\end{document}